\documentclass[11pt,a4paper]{article}
\usepackage[hyperref]{acl2020}
\usepackage{times}
\usepackage{latexsym}

\usepackage{amsmath}
\usepackage{amsfonts}
\usepackage{url}
\usepackage{booktabs}
\usepackage{pifont}
\usepackage{multirow}
\usepackage{amsthm}
\usepackage{graphics}
\usepackage{tikz-dependency}
\usepackage{diagbox}
\usepackage{wrapfig}
\usepackage{enumitem}
\usepackage{makecell}
\usepackage{array}
\usepackage{subcaption}
\usepackage{xspace}
\usepackage{afterpage}
\usepackage{float}
\usepackage[belowskip=5pt,aboveskip=5pt]{caption}
\usepackage{titlesec}
\usepackage{tablefootnote}

\usepackage{microtype}

\aclfinalcopy 
\def\aclpaperid{1067} 

\makeatletter
\newenvironment{smalleralign}[1][\small]
 {\par\nopagebreak\leavevmode\vspace*{-\baselineskip}%
  \skip0=\abovedisplayskip
  #1%
  \def\maketag@@@##1{\hbox{\m@th\normalfont\normalsize##1}}%
  \abovedisplayskip=\skip0
  \align}
 {\endalign\ignorespacesafterend}
\makeatother

\newcolumntype{H}{>{\setbox0=\hbox\bgroup}c<{\egroup}@{}}
\newcolumntype{C}{@{\extracolsep{5pt}}c@{\extracolsep{}}}
\newcommand\ul[1]{\underline{\smash{#1}}}
\newcommand\tline[2]{$\underset{\text{#1}}{\text{\underline{#2}}}$}
\newcommand\tlinesmash[2]{$\underset{\text{#1}}{\text{\ul{#2}}}$}

\newcommand{\cmark}{\ding{51}}
\newcommand{\xmark}{\textcolor{lightgray}{\ding{55}}}

\newcommand{\better}[1]{\textcolor{blue}{\textbf{#1}}$\uparrow$}
\newcommand{\worse}[1]{\textcolor{red}{#1}$\downarrow$}

\newcommand{\MLP}{\text{MLP}}
\newcommand{\softmax}{\mathrm{softmax}}

\def\modelname{SpanRel\xspace}
\def\benchmarkname{GLAD\xspace}

\def\tightcol{\hskip 2pt}
\def\smallcol{\hskip 5pt}

\title{Generalizing Natural Language Analysis through \\ Span-relation Representations}

\author{
Zhengbao Jiang$^1$, Wei Xu$^2$, Jun Araki$^3$, Graham Neubig$^1$ \\
Language Technologies Institute, Carnegie Mellon University$^1$ \\
Department of Computer Science and Engineering, Ohio State University$^2$ \\
Bosch Research North America$^3$ \\
\texttt{\{zhengbaj,gneubig\}@cs.cmu.edu}$^1$ \\
\texttt{xu.1265@osu.edu$^2$, jun.araki@us.bosch.com$^3$}
}

\date{}

\begin{document}
\maketitle
\begin{abstract}
Natural language processing covers a wide variety of tasks predicting syntax, semantics, and information content, and usually each type of output is generated with specially designed architectures.
In this paper, we provide the simple insight that a great variety of tasks can be represented in a single unified format consisting of labeling spans and relations between spans, thus a single task-independent model can be used across different tasks.
We perform extensive experiments to test this insight on 10 disparate tasks spanning dependency parsing (syntax), semantic role labeling (semantics), relation extraction (information content), aspect based sentiment analysis (sentiment), and many others, achieving performance comparable to state-of-the-art specialized models.
We further demonstrate benefits of multi-task learning, and also show that the proposed method makes it easy to analyze differences and similarities in how the model handles different tasks.
Finally, we convert these datasets into a unified format to build a benchmark, which provides a holistic testbed for evaluating future models for generalized natural language analysis.
\end{abstract}

\section{Introduction}
A large number of natural language processing (NLP) tasks exist to analyze various aspects of human language, including syntax (e.g., constituency and dependency parsing), semantics (e.g., semantic role labeling), information content (e.g., named entity recognition and relation extraction), or sentiment (e.g., sentiment analysis).
At first glance, these tasks are seemingly very different in both the structure of their output and the variety of information that they try to capture.
To handle these different characteristics, researchers usually use specially designed neural network architectures.
In this paper we ask the simple questions:
are the task-specific architectures really necessary?
Or with the appropriate representational methodology, can we devise \emph{a single model that can perform --- and achieve state-of-the-art performance on --- a large number of natural language analysis tasks}?

\begin{figure}
\begin{center}
\includegraphics[width=0.7\columnwidth]{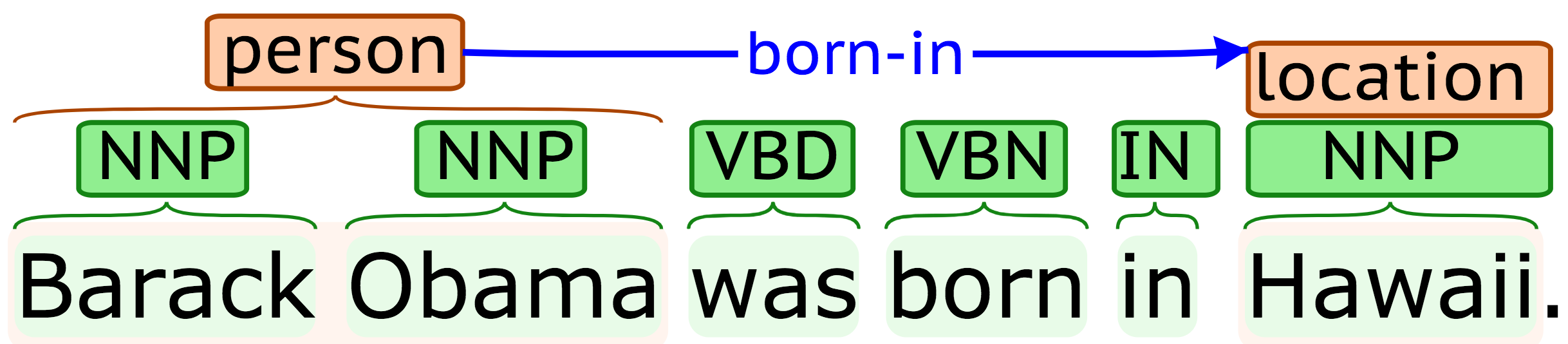}
\end{center}
\caption{An example from BRAT, consisting of \textcolor{green}{POS}, \textcolor{orange}{NER}, and \textcolor{blue}{RE}.}
\label{fig:brat}
\end{figure}

\begin{table*}[tb]
\resizebox{\textwidth}{!}{%
\begin{tabular}{ccccccccccc}
\toprule
 & \multicolumn{4}{c}{\textbf{Information Extraction}} & \multirow{2}{*}{\textbf{POS}} & \multicolumn{2}{c}{\textbf{Parsing}} & \multirow{2}{*}{\textbf{SRL}} & \multicolumn{2}{c}{\textbf{Sentiment}} \\
\cline{2-5} 
\cline{7-8}
\cline{10-11}
 & NER & RE & Coref. & OpenIE  & & Dep. & Consti. & & ABSA & ORL \\
\toprule
\multicolumn{11}{c}{Different Models for Different Tasks} \\
\midrule
ELMo \citep{peters-etal-2018-deep} & \cmark & \xmark & \cmark & \xmark & \xmark & \xmark & \xmark & \xmark & \cmark & \xmark \\
BERT \citep{devlin-etal-2019-bert} & \cmark & \xmark & \xmark & \xmark & \xmark & \xmark & \xmark & \xmark & \xmark & \xmark \\
SpanBERT \citep{joshi:19:spanbert} & \xmark & \cmark & \cmark & \xmark & \xmark & \xmark & \xmark & \xmark & \xmark & \xmark \\
\midrule
\multicolumn{11}{c}{Single Model for Different Tasks} \\
\midrule
\citet{guo-etal-2016-unified} & \xmark & \cmark & \xmark & \xmark & \xmark & \xmark & \xmark & \cmark & \xmark & \xmark \\
\citet{swayamdipta-etal-2018-syntactic} & \xmark & \xmark & \cmark & \xmark & \xmark & \xmark & \cmark & \cmark & \xmark & \xmark \\
\citet{strubell-etal-2018-linguistically} & \xmark & \xmark & \xmark & \xmark & \cmark & \cmark & \xmark & \cmark & \xmark & \xmark \\
\citet{clark-etal-2018-semi} & \cmark & \xmark & \xmark & \xmark & \cmark & \cmark & \xmark & \xmark & \xmark & \xmark \\
\citet{luan:18:mtlie,luan-etal-2019-general} & \cmark & \cmark & \cmark & \xmark & \xmark & \xmark & \xmark & \xmark & \xmark & \xmark \\
\citet{dixit-al-onaizan-2019-span} & \cmark & \cmark & \xmark & \xmark & \xmark & \xmark & \xmark & \xmark & \xmark & \xmark \\
\citet{marasovic-frank-2018-srl4orl} & \xmark & \xmark & \xmark & \xmark & \xmark & \xmark & \xmark & \cmark & \xmark & \cmark \\
\citet{hashimoto:17:mtlsocher} & \xmark & \xmark & \xmark & \xmark & \cmark & \cmark & \xmark & \xmark & \xmark & \xmark \\
 This Work & \cmark & \cmark & \cmark & \cmark & \cmark & \cmark & \cmark & \cmark & \cmark & \cmark \\
\bottomrule
\end{tabular}
}
\caption{A comparison of the tasks covered by previous work and our work.
}
\vspace{-3mm}
\label{tab:related_work}
\end{table*}

Interestingly, in the domain of \emph{efficient human annotation interfaces}, it is already standard to use unified representations for a wide variety of NLP tasks.
\autoref{fig:brat} shows one example of the BRAT \citep{stenetorp-etal-2012-brat} annotation interface, which has been used for annotating data for tasks as broad as part-of-speech tagging, named entity recognition, relation extraction, and many others.
Notably, this interface has a single unified format that consists of spans (e.g., the span of an entity), labels on the spans (e.g., the variety of entity such as ``person'' or ``location''), and labeled relations between the spans (e.g., ``born-in'').
These labeled relations can form a tree or a graph structure, expressing the linguistic structure of sentences (e.g., dependency tree). 
We detail this BRAT format and how it can be used to represent a wide number of natural language analysis tasks in \autoref{sec:brat}. 

The simple hypothesis behind our paper is: \emph{if humans can perform natural language analysis in a single unified format, then perhaps machines can as well}.
Fortunately, there already exist NLP models that perform span prediction and prediction of relations between pairs of spans, such as the end-to-end coreference model of \citet{lee:17:e2ecoref}.
We extend this model with minor architectural modifications (which are \emph{not} our core contributions) and pre-trained contextualized representations (e.g., BERT; \citet{devlin-etal-2019-bert}\footnote{In contrast to work on pre-trained contextualized representations like ELMo \citep{peters-etal-2018-deep} or BERT \citep{devlin-etal-2019-bert} that learn unified \emph{features} to represent the \emph{input} in different tasks, we propose a unified \emph{representational methodology} that represents the \emph{output} of different tasks. Analysis models using BERT still use special-purpose output predictors for specific tasks or task classes.}) then demonstrate the applicability and versatility of this single model on 10 tasks, including named entity recognition (NER), relation extraction (RE), coreference resolution (Coref.), open information extraction (OpenIE), part-of-speech tagging (POS), dependency parsing (Dep.), constituency parsing (Consti.), semantic role labeling (SRL), aspect based sentiment analysis (ABSA), and opinion role labeling (ORL).
While previous work has used similar formalisms to \emph{understand} the representations learned by pre-trained embeddings \citep{tenney:19:bertpipeline,tenney:19:bertprobe}, to the best of our knowledge this is the first work that uses such a unified model to actually \emph{perform analysis}.
Moreover, we demonstrate that despite the model's simplicity, it can achieve comparable performance with special-purpose state-of-the-art models on the tasks above (\autoref{tab:related_work}).
We also demonstrate that this framework allows us to easily perform multi-task learning (MTL), leading to improvements when there are related tasks to be learned from or data is sparse.
Further analysis shows that dissimilar tasks exhibit divergent attention patterns, which explains why MTL is harmful on certain tasks.
We have released our code and the \textbf{G}eneral \textbf{L}anguage \textbf{A}nalysis \textbf{D}atasets (\benchmarkname) benchmark with 8 datasets covering 10 tasks in the BRAT format at \url{https://github.com/neulab/cmu-multinlp},
and provide a leaderboard to facilitate future work on generalized models for NLP.

\begin{table*}[tb]
\begin{subtable}[t]{0.47\textwidth}
\renewcommand\thesubtable{Table \thetable a}
\renewcommand{\arraystretch}{1.35}
\centering
\resizebox{\columnwidth}{!}{
\begin{tabular}[t]{ll}
\toprule
 \textbf{Task} & \textbf{\ul{Spans} annotated with labels} \\
\hline \\[\dimexpr-\normalbaselineskip+6pt]
NER & \tlinesmash{person}{Barack Obama} was born in \tlinesmash{location}{Hawaii}. \\[16pt]
Consti. & \tline{S}{And \tline{NP}{\tlinesmash{NP}{their suspicions} \tline{PP}{of \tlinesmash{NP}{each other}}} \tline{VP}{run \tlinesmash{ADVP}{deep}}.} \vspace{8pt} \\[16pt]
 
 POS & \tlinesmash{WP}{What} \tlinesmash{NN}{kind} \tlinesmash{IN}{of} \tlinesmash{NN}{memory}? \\[16pt]
 ABSA & Great laptop that offers many great \tlinesmash{positive}{features}! \\[13pt]
\bottomrule
\end{tabular}
}
\renewcommand{\arraystretch}{1.0}
\subcaption{Span-oriented tasks. Spans are annotated by underlines and their labels.}
\vspace{-3mm}
\label{tab:demo_span}
\end{subtable}
\hspace{\fill}
\begin{subtable}[t]{0.53\textwidth}
\renewcommand\thesubtable{Table \thetable b}
\centering
\resizebox{\columnwidth}{!}{
\begin{tabular}[t]{Hll}
\toprule
\textbf{Category} & \textbf{Task} & \textbf{\ul{Spans} and \begin{tikzpicture}[baseline=(rel.base)]\node(rel)[draw,rounded corners]{relations};\end{tikzpicture} annotated with labels} \\
\midrule
\multirow{10}{*}{Relation} & \raisebox{1\height}{RE} & 
\begin{dependency}
\begin{deptext}
The \& \ul{burst} \& has been caused by \& \ul{pressure}.\\
\end{deptext}
\depedge[edge height=0.8ex]{4}{2}{cause-effect}
\end{dependency} 
\vspace{-5pt} \\
 & \raisebox{1\height}{Coref.} & 
\begin{dependency}
\begin{deptext}
I voted for \& \ul{Tom} \& because \& \ul{he} \& is clever.\\
\end{deptext}
\depedge[edge height=0.8ex]{4}{2}{coref.}
\end{dependency}  
\vspace{-5pt} \\
 & \raisebox{1\height}{SRL} & 
\begin{dependency}
\begin{deptext}
\ul{We} \& \quad \ul{brought} \& \ul{you} \& \ul{the tale of two cities}.\\
\end{deptext}
\depedge[edge height=0.8ex]{2}{1}{ARG0}
\depedge[edge height=0.8ex]{2}{3}{ARG2}
\depedge[edge height=2.3ex]{2}{4}{ARG1}
\end{dependency}
\vspace{-5pt} \\
 & \raisebox{1\height}{OpenIE} & 
\begin{dependency}
\begin{deptext}
\ul{The four lawyers} \& \ul{climbed out} \& \ul{from under a table}.\\ 
\end{deptext}
\depedge[edge height=0.8ex]{2}{1}{ARG0}
\depedge[edge height=0.8ex]{2}{3}{ARG1}
\end{dependency} 
\vspace{-5pt} \\
 & \raisebox{1\height}{Dep.} & 
\begin{dependency}
\begin{deptext}
\ul{The} \& \ul{entire} \& \ul{division} \& \ul{employs} \& \ul{about} \& \ul{850} \& \ul{workers}.\\
\end{deptext}
\depedge[edge height=4ex]{3}{1}{det}
\depedge[edge height=1.5ex]{3}{2}{amod}
\depedge[edge height=1.5ex]{4}{3}{nsubj}
\depedge[edge height=1.5ex]{6}{5}{advmod}
\depedge[edge height=1.5ex]{7}{6}{nummod}
\depedge[edge height=4ex]{4}{7}{dobj}
\end{dependency}
\vspace{-5pt} \\
 & \raisebox{1\height}{ORL} & 
\begin{dependency}
\begin{deptext}
\ul{We} \& therefore as MDC \& \ul{do not accept} \& \ul{this result}.\\
\end{deptext}
\depedge[edge height=0.8ex]{3}{1}{holder}
\depedge[edge height=0.8ex]{3}{4}{target}
\end{dependency}
\vspace{-5pt} \\
\bottomrule
\end{tabular}
}
\subcaption{Relation-oriented tasks. Directed arcs indicate the relations between spans.}
\vspace{-3mm}
\label{tab:demo_rel}
\end{subtable}
\end{table*}

\section{Span-relation Representations}\label{sec:brat}
In this section, we explain how the BRAT format can be used to represent a large number of tasks.
There are two fundamental types of annotations: span annotations and relation annotations.
Given a sentence $\mathbf{x}=[w_1, w_2, ..., w_n]$ of $n$ tokens, a span annotation $(s_i, l_i)$ consists of a contiguous span of tokens $s_i=[w_{b_i}, w_{b_i+1}, ..., w_{e_i}]$ and its label $l_i$ ($l_i\in\mathcal{L}$), where $b_i$/$e_i$ are the start/end indices respectively, and $\mathcal{L}$ is a set of span labels.
A relation annotation $(s_j, s_k, r_{jk})$ refers to a relation $r_{jk}$ ($r_{jk}\in\mathcal{R}$) between the head span $s_j$ and the tail span $s_k$, where $\mathcal{R}$ is a set of relation types.
This span-relation representation can easily express many tasks by defining $\mathcal{L}$ and $\mathcal{R}$ accordingly, as summarized in \subref{tab:demo_span} and \subref{tab:demo_rel}.
These tasks fall in two categories: \textbf{span-oriented tasks}, where the goal is to predict labeled spans (e.g., named entities in NER) and \textbf{relation-oriented tasks}, where the goal is to predict relations between two spans (e.g., relation between two entities in RE).
For example, constituency parsing \citep{collins-1997-three} is a span-oriented task aiming to produce a syntactic parse tree for a sentence, where each node of the tree is an individual span associated with a constituent label.
Coreference resolution \citep{pradhan:12:conll2012} is a relation-oriented task that links an expression to its mentions within or beyond a single sentence.
Dependency parsing \citep{kubler2009dependency} is also a relation-oriented task that aims to relate a word (single-word span) to its syntactic parent word with the corresponding dependency type.
Detailed explanations of all tasks can be found in \autoref{ap:tasks}.

While the tasks above represent a remarkably broad swath of NLP, it is worth mentioning what we have \emph{not} covered, to properly scope this work.
Notably, sentence-level tasks such as text classification and natural language inference are not covered, although they can also be formulated using this span-relation representation by treating the entire sentence as a span. We chose to omit these tasks because they are already well-represented by previous work on generalized architectures \citep{lan-xu-2018-neural} and multi-task learning \citep{devlin-etal-2019-bert,liu:19:mtdnn}, and thus we mainly focus on tasks using phrase-like spans.
In addition, the span-relation representations described here are designed for natural language \emph{analysis}, and cannot handle tasks that require \emph{generation} of text, such as machine translation \citep{bojar:14:wmt14}, dialog response generation \citep{lowe:15:ubuntu}, and summarization \citep{nallapati:16:sum}.
There are also a small number of analysis tasks such as semantic parsing to logical forms \citep{banarescu:13:amr} where the outputs are not directly associated with spans in the input, and handling these tasks is beyond the scope of this work.

\section{Span-relation Model}\label{sec:spanrelationmodel}
Now that it is clear that a very large number of analysis tasks can be formulated in a single format, we turn to devising a single model that can solve these tasks.
We base our model on a span-based model first designed for end-to-end coreference resolution \citep{lee:17:e2ecoref}, which is then adapted for other tasks \citep{he-etal-2018-jointly,luan:18:mtlie,luan-etal-2019-general,dixit-al-onaizan-2019-span,zhan:19:spanoie}.
At the core of the model is a module to represent each span as a fixed-length vector, which is used to predict labels for spans or span pairs.
We first briefly describe the span representation used and proven to be effective in previous works, then highlight some details we introduce to make this model generalize to a wide variety of tasks.

\paragraph{Span Representation}
Given a sentence $\mathbf{x}=[w_1, w_2, ..., w_n]$ of $n$ tokens, a span $s_i=[w_{b_i}, w_{b_i+1}, ..., w_{e_i}]$ is represented by concatenating two components: a \emph{content representation} $\mathbf{z}^c_i$ calculated as the weighted average across all token embeddings in the span, and a \emph{boundary representation} $\mathbf{z}^u_i$ that concatenates the embeddings at the start and end positions of the span. Specifically,
\begin{smalleralign}[\small]
\mathbf{c}_1,\mathbf{c}_2,...,\mathbf{c}_n &= \text{TokenRepr}(w_1, w_2, ..., w_n), \label{eq:tokenrepr} \\
\mathbf{u}_1,\mathbf{u}_2,...,\mathbf{u}_n &= \text{BiLSTM}(\mathbf{c}_1,\mathbf{c}_2,...,\mathbf{c}_n), \label{eq:bilstm} \\
\mathbf{z}_i^c &= \text{SelfAttn}(\mathbf{c}_{b_i},\mathbf{c}_{b_i+1},...,\mathbf{c}_{e_i}), \\
\mathbf{z}_i^u &= [\mathbf{u}_{b_i};\mathbf{u}_{e_i}], \mathbf{z}_i = [\mathbf{z}_i^c;\mathbf{z}_i^u], 
\end{smalleralign}
where $\text{TokenRepr}$ could be non-contextualized, such as GloVe \citep{pennington:14:glove}, or contextualized, such as BERT \citep{devlin-etal-2019-bert}. We refer to \citet{lee:17:e2ecoref} for further details.

\begin{table*}[tb]
\small
\centering
\resizebox{\textwidth}{!}{%
\begin{tabular}{ccccrrc}
\toprule
\textbf{Dataset} & \textbf{Domain} & \textbf{\#Sent.} & \textbf{Task} & \textbf{\#Spans} & \textbf{\#Relations} & \textbf{Metric} \\
\midrule
\multirow{1}{*}{Wet Lab Protocols} & \multirow{2}{*}{biology} & \multirow{2}{*}{14,301} & NER & 60,745 & - & F$_1$ \\
\multirow{1}{*}{\citep{kulkarni-etal-2018-annotated}} & & & RE & 60,745 & 43,773 & F$_1$ \\
\midrule
\multirow{1}{*}{CoNLL-2003 \citep{sang:03:conll2003}} & \multirow{1}{*}{news} & \multirow{1}{*}{20,744} & \multirow{1}{*}{NER} & \multirow{1}{*}{35,089} & \multirow{1}{*}{-} & \multirow{1}{*}{F$_1$} \\
\midrule
SemEval-2010 Task 8 \citep{hendrickx:10:semeval2010} & \multirow{1}{*}{misc.} & \multirow{1}{*}{10,717} & \multirow{1}{*}{RE} & \multirow{1}{*}{21,437} & \multirow{1}{*}{10,717} & \multirow{1}{*}{Macro F$_1$ $^\circ$} \\
\midrule
\multirow{5}{*}{\shortstack{OntoNotes 5.0 $^\star$\\\citep{pradhan:13:ontonotes}}} & \multirow{5}{*}{misc.} & \multirow{5}{*}{94,268} & Coref. & 194,477 & 1,166,513 & Avg F$_1$ \\
& & & SRL & 745,796 & 543,534 & F$_1$ \\
& & & POS & 1,631,995 & - & Accuracy \\
& & & Dep. & 1,722,571 & 1,628,558 & LAS \\
& & & Consti. & 1,320,702 & - & Evalb F$_1$ $^\dagger$ \\
\midrule
\multirow{3}{*}{\shortstack{Penn Treebank\\\citep{marcus:94:ptb}}} & \multirow{3}{*}{speech, news} & 49,208 & POS & 1,173,766 & - & Accuracy \\
& & 43,948 & Dep. & 1,090,777 & 1,046,829 & LAS \\
& & 43,948 & Consti. & 871,264 & - & Evalb F$_1$ $^\dagger$ \\
\midrule
OIE2016 \citep{stanovsky-dagan-2016-creating} & \multirow{1}{*}{news, Wiki} & \multirow{1}{*}{2,534} & \multirow{1}{*}{OpenIE} & \multirow{1}{*}{15,717} & \multirow{1}{*}{12,451} & \multirow{1}{*}{F$_1$} \\
\midrule
MPQA 3.0 \citep{deng:15:mpqa} & \multirow{1}{*}{news} & \multirow{1}{*}{3,585} & \multirow{1}{*}{ORL} & \multirow{1}{*}{13,841} & \multirow{1}{*}{9,286} & \multirow{1}{*}{F$_1$} \\
\midrule
SemEval-2014 Task 4 \citep{pontiki:14:semeval2014} & \multirow{1}{*}{reviews} & \multirow{1}{*}{4,451} & \multirow{1}{*}{ABSA} & \multirow{1}{*}{7,674} & \multirow{1}{*}{-} & \multirow{1}{*}{Accuracy $^\circ$} \\
\bottomrule
\end{tabular}
}
\caption{
Statistics of \benchmarkname, consisting of 10 tasks from 8 datasets. 
$^\star$ Following \citet{he-etal-2018-jointly}, we use a subset of OntoNotes 5.0 dataset based on CoNLL 2012 splits \citep{pradhan:12:conll2012}.
$^\circ$ Previous works use gold standard spans in these evaluations.
$^\dagger$ We use the bracket scoring program Evalb \citep{collins-1997-three} in constituency parsing.
}
\vspace{-3mm}
\label{tab:dataset}
\end{table*}

\paragraph{Span and Relation Label Prediction}
Since we extract spans and relations in an \emph{end-to-end} fashion, we introduce two additional labels \texttt{NEG\_SPAN} and \texttt{NEG\_REL} in $\mathcal{L}$ and $\mathcal{R}$ respectively. \texttt{NEG\_SPAN} indicates invalid spans (e.g., spans that are not named entities in NER) and \texttt{NEG\_REL} indicates invalid span pairs without any relation between them (i.e., no relation exists between two arguments in SRL).
We first predict labels for \emph{all} spans up to a length of $l$ words using a multilayer perceptron (MLP): $\softmax(\MLP^\text{span}(\mathbf{z}_i)) \in \Delta^{|\mathcal{L}|}$, where $\Delta^{|\mathcal{L}|}$ is a $|\mathcal{L}|$-dimensional simplex.
Then we keep the top $K = \tau \cdot n$ spans with the lowest \texttt{NEG\_SPAN} probability in relation prediction for efficiency, where smaller pruning threshold $\tau$ indicates more aggressive pruning.
Another MLP is applied to pairs of the remaining spans to produce their relation scores: $\mathbf{o}_{jk}=\MLP^\text{rel}([\mathbf{z}_j;\mathbf{z}_k;\mathbf{z}_j\cdot \mathbf{z}_k]) \in \mathbb{R}^{|\mathcal{R}|}$, where $j$ and $k$ index two spans.

\paragraph{Application to Disparate Tasks}
For most of the tasks, we can simply maximize the probability of the ground truth relation for \emph{all pairs of the remaining spans}.
However, some tasks might have different requirements, e.g., coreference resolution aims to cluster spans referring to the same concept and we do not care about which antecedent a span is linked to if there are multiple ones. Thus, we provide two training loss functions:
\begin{enumerate}[leftmargin=15pt]
\item \textbf{Pairwise} Maximize the probabilities of the ground truth relations for all pairs of the remaining spans independently: $\softmax(\mathbf{o}_{jk})_{r_{jk}}$, where $r_{jk}$ indexes the ground truth relation.
\item \textbf{Head}
Maximize the probability of ground truth head spans for a specific span $s_j$: $\sum_{k \in \text{head}(s_j)}{\softmax([o_{j1}, o_{j2}, ..., o_{jK}])_k}$,  where $\text{head}(\cdot)$ returns indices of one or more heads and $o_{j\cdot}$ is the corresponding scalar from $\mathbf{o}_{j\cdot}$ indicating how likely two spans are related.
\end{enumerate}
We use option 1 for all tasks except for coreference resolution which uses option 2.
Note that the above loss functions \emph{only} differ in how relation scores are normalized and the other parts of the model remain the same across different tasks.
At test time, we follow previous inference methods to generate valid outputs.
For coreference resolution, we link a span to the antecedent with highest score \citep{lee:17:e2ecoref}.
For constituency parsing, we use greedy top-down decoding to generate a valid parse tree \citep{stern:17:spanconsti}.
For dependency parsing, each word is linked to exactly one parent with the highest relation probability.
For other tasks, we predict relations for all span pairs and use those not predicted as \texttt{NEG\_REL} to construct outputs.

Our core insight is that the above formulation is largely \emph{task-agnostic}, meaning that a task can be modeled in this framework as long as it can be formulated as a span-relation prediction problem with properly defined span labels $\mathcal{L}$ and relation labels $\mathcal{R}$.
As shown in \autoref{tab:related_work}, this unified \textbf{Span}-\textbf{Rel}ation (\modelname) model makes it simple to scale to a large number of language analysis tasks, with breadth far beyond that of previous work.

\begin{table*}[tb]
\centering
\resizebox{\textwidth}{!}{
\begin{tabular}{ccccll|cc}
\toprule
\textbf{Category} & \textbf{Task} & \textbf{Metric} & \textbf{Dataset} & \textbf{Setting} & \textbf{SOTA Model} & \textbf{Previous SOTA} & \textbf{Our Model} \\
\midrule
\multirow{6}{*}{IE} & \multirow{2}{*}{NER} & \multirow{2}{*}{F$_1$} & CoNLL03 & BERT & \citet{devlin-etal-2019-bert} & 92.8 & 92.2 \\
 & & & WLP & ELMo & \citet{luan-etal-2019-general} & 79.5 & 79.2 \\
 \cmidrule{2-8}
 & \multirow{2}{*}{RE} & Macro F1 & SemEval10 & BERT, gold & \citet{wu:19:bertrc} & 89.3 & 87.4 \\
 & & F$_1$ & WLP & ELMo & \citet{luan-etal-2019-general} & 64.1 & 65.5 \\
  \cmidrule{2-8}
 & \multirow{1}{*}{Coref.} & Avg F$_1$ & OntoNotes & GloVe, CharCNN & \citet{lee:17:e2ecoref}$^\circ$ & 62.0 & 61.1 \\
  \cmidrule{2-8}
 & \multirow{1}{*}{OpenIE} & F$_1$ & OIE2016 & ELMo & \citet{stanovsky-etal-2018-supervised}$^\star$ & 31.1 & 35.2 \\
\midrule
\multicolumn{2}{c}{SRL} & F$_1$ & OntoNotes & ELMo & \citet{he-etal-2018-jointly}$^\dagger$ & 82.9 & 82.4 \\
\midrule
\multirow{2}{*}{Parsing} & Dep. & LAS & PTB & ELMo & \citet{clark-etal-2018-semi} & 94.4 & 94.7 \\
  \cmidrule{2-8}
 & Consti. & Evalb F$_1$ & PTB & BERT & \citet{kitaev-etal-2019-multilingual} & 95.6 & 95.5 \\
\midrule
\multirow{2}{*}{Sentiment} & ABSA & Accuracy & SemEval14 & BERT, gold & \citet{xu-etal-2019-bert}$^\triangleleft$ & 85.0/78.1 & 85.5/76.6 \\
 \cmidrule{2-8}
 & ORL & F$_1$ & MPQA 3.0 & GloVe, gold & \citet{marasovic-frank-2018-srl4orl}$^\star$ & 56.4 & 55.6 \\
\midrule
\multicolumn{2}{c}{POS} & Accuracy & PTB & ELMo & \citet{clark-etal-2018-semi} & 97.7 & 97.7 \\
\bottomrule
\end{tabular}
}
\caption{Comparison between \modelname models and task-specific SOTA models.\protect\footnotemark{}
Following \citet{luan-etal-2019-general}, we perform NER and RE jointly on WLP dataset.
We use gold entities in SemEval-2010 Task 8, gold aspect terms in SemEval-2014 Task 4, and gold opinion expressions in MPQA 3.0 to be consistent with existing works.
}
\vspace{-3mm}
\label{tab:sota}
\end{table*}
\afterpage{
\footnotetext{
$^\circ$ The small version of \citet{lee:17:e2ecoref}'s method with 100 antecedents and no speaker features.
$^\star$ For OpenIE and ORL, we use span-based F$_1$ instead of syntactic-head-based F$_1$ and binary coverage F$_1$ used in the original papers because they are biased towards extracting long spans.
$^\dagger$ For SRL, we choose to compare with \citet{he-etal-2018-jointly} because they also extract predicates and arguments in an end-to-end way.
$^\triangleleft$ We follow \citet{xu-etal-2019-bert} to report accuracy of restaurant and laptop domain separately in ABSA.
}
}

\paragraph{Multi-task Learning}
The \modelname model makes it easy to perform multi-task learning (MTL) by sharing all parameters except for the MLPs used for label prediction.
However, because different tasks capture different linguistic aspects, they are not equally beneficial to each other.
It is expected that jointly training on related tasks is helpful, while forcing the same model to solve unrelated tasks might even hurt the performance \citep{ruder:17:mtl}.
Compared to manually choosing source tasks based on prior knowledge, which might be sub-optimal when the number of tasks is large, \modelname offers a systematic way to examine relative benefits of source-target task pairs by either performing pairwise MTL or attention-based analysis, as we will show in \autoref{sec:exp_pair_mtl}.

\section{\benchmarkname Benchmark and Results}

We first describe our \textbf{G}eneral \textbf{L}anguage \textbf{A}nalysis \textbf{D}atasets (\benchmarkname) benchmark and evaluation metrics, then conduct experiments to (1) verify that \modelname can achieve comparable performance across all tasks (\autoref{sec:exp_sota}), and (2) demonstrate its benefits in multi-task learning (\autoref{sec:exp_mtl}).

\subsection{Experimental Settings}
\label{sec:datasets}
\paragraph{\benchmarkname Benchmark and Evaluation Metrics} As summarized in \autoref{tab:dataset}, we convert 8 widely used datasets with annotations of 10 tasks into the BRAT format and include them in the \benchmarkname benchmark. It covers diverse domains, providing a holistic testbed for natural language analysis evaluation.
The major evaluation metric is span-based F$_1$ (denoted as F$_1$), a standard metric for SRL.
Precision is the proportion of extracted spans (spans not predicted as \texttt{NEG\_SPAN}) that are consistent with the ground truth.
Recall is the proportion of ground truth spans that are correctly extracted.
Span F$_1$ is also applicable to relations, where an extracted relation (relations not predicted as \texttt{NEG\_REL}) is correct iff both head and tail spans have correct boundaries and the predicted relation is correct.
To make fair comparisons with existing works, we also compute standard metrics for different tasks, as listed in \autoref{tab:dataset}.

\paragraph{Implementation Details}
We attempted four token representation methods (\autoref{eq:tokenrepr}), namely GloVe \citep{pennington:14:glove}, ELMo \citep{peters-etal-2018-deep}, BERT \citep{devlin-etal-2019-bert}, and SpanBERT \citep{joshi:19:spanbert}.
We use BERT$_\text{base}$ in our main results and report BERT$_\text{large}$ in \autoref{ap:bertlarge}.
A three-layer BiLSTM with 256 hidden units is used (\autoref{eq:bilstm}).
Both span and relation prediction MLPs have two layers with 128 hidden units.
Dropout \citep{srivastava:14:dropout} of 0.5 is applied to all layers.
For GloVe and ELMo, we use Adam \citep{kingma:14:adam} with learning rate of 1e-3 and early stop with patience of 3.
For BERT and SpanBERT, we follow standard fine-tuning with learning rate of 5e-5, $\beta_1=0.9$, $\beta_2=0.999$, L2 weight decay of 0.01, warmup over the first 10\% steps, and number of epochs tuned on development set.
Task-specific hyperparameters maximal span length and pruning ratio are tuned on development set and listed in \autoref{ap:hyper}.

\subsection{Comparison with Task-specific SOTA}\label{sec:exp_sota}

We compare the \modelname model with state-of-the-art task-specific models by training on data from a single task.
By doing so we attempt to answer the research question ``can a single model with minimal task-specific engineering achieve competitive or superior performance to other models that have been specifically engineered?''
We select competitive SOTA models mainly based on settings, e.g., single-task learning and end-to-end extraction of spans and relations.
To make fair comparisons, token embeddings (GloVe, ELMo, BERT) and other hyperparameters (e.g., the number of antecedents in Coref. and the maximal span length in SRL) in our method are set to match those used by SOTA models, to focus on differences brought about by the model architecture.

As shown in \autoref{tab:sota}, the \modelname model achieves comparable performances as task-specific SOTA methods (regardless of whether the token representation is contextualized or not).
This indicates that the span-relation format can generically represent a large number of natural language analysis tasks and it is possible to devise a single unified model that achieves strong performance on all of them.
It provides a strong and generic baseline for natural language analysis tasks and a way to examine the usefulness of task-specific designs.

\subsection{Multi-task Learning with \modelname}\label{sec:exp_mtl}

To demonstrate the benefit of the \modelname model in MTL, we perform single-task learning (STL) and MTL across all tasks using end-to-end settings.\footnote{Span-based F$_1$ is used as the evaluation metric in SemEval-2010 Task 8 and SemEval-2014 Task 4 as opposed to macro F$_1$ and accuracy reported in the original papers because we aim at end-to-end extractions.}
Following \citet{liu:19:mtdnn}, we perform MTL+fine-tuning and show the results in separate columns of \autoref{tab:mtl}.
Contextualized token representations yield significantly better results than GloVe on all tasks, indicating that pre-training on large corpora is almost universally helpful to NLP tasks.
Comparing the results of MTL+fine-tuning with STL, we found that performance with GloVe drops on 8 out of 15 tasks, most of which are tasks with relatively sparse data.
It is probably because the capacity of the GloVe-based model is too small to store all the patterns required by different tasks.
The results of contextualized representations are mixed, with some tasks being improved and others remaining the same or degrading.
We hypothesize that this is because different tasks capture different linguistic aspects, thus are not equally helpful to each other.
Reconciling these seemingly different tasks in the same model might be harmful to some tasks.
Notably, as the contextualized representations become stronger, the performance of MTL+FT becomes more favorable.
5 out of 15 tasks (NER, RE, OpenIE, SRL, ORL) observe statistically significant improvements (p-value $<0.05$ with paired bootstrap re-sampling) with SpanBERT, a contextualized embedding pre-trained with span-based training objectives, while only one task degrades (ABSA), indicating its superiority in reconciling spans from different tasks.
The \benchmarkname benchmark provides a holistic testbed for evaluating natural language analysis capability.

\paragraph{Task Relatedness Analysis}\label{sec:exp_pair_mtl}
To further investigate how different tasks interact with each other, we choose five source tasks (i.e., tasks used to improve other tasks, e.g., POS, NER, Consti., Dep., and SRL) that have been widely used in MTL \citep{hashimoto:17:mtlsocher,strubell-etal-2018-linguistically} and six target tasks (i.e., tasks to be improved, e.g., OpenIE, NER, RE, ABSA, ORL, and SRL) to perform pairwise multi-task learning.

\begin{figure*}[tb]
\begin{minipage}{\textwidth}
\begin{minipage}[b]{0.78\textwidth}
\centering
\resizebox{1.0\columnwidth}{!}{
\begin{tabular}{c|c|c@{\smallcol}c@{\smallcol}c@{\smallcol}c@{\smallcol}c||c|c@{\smallcol}c@{\smallcol}c@{\smallcol}c@{\smallcol}c@{\smallcol}c@{\smallcol}c}
\toprule
 & \multicolumn{6}{c||}{\textbf{GloVe}} & \multicolumn{6}{c}{\textbf{BERT$_\text{base}$}} \\
\midrule
\backslashbox{\textbf{Target}}{\textbf{Source}} & \textbf{STL} & \textbf{POS} & \textbf{NER} & \textbf{Consti.} & \textbf{Dep.} & \textbf{SRL} & \textbf{STL} & \textbf{POS} & \textbf{NER} & \textbf{Consti.} & \textbf{Dep.} & \textbf{SRL} \\
\midrule
OpenIE & 28.3 & \better{29.9} & \worse{27.0} & \better{31.2} & \better{32.9} & \better{34.1} & 36.7 & \worse{34.0} & \worse{34.3} & \worse{35.2} & \better{37.8} & \better{38.3} \\
NER (WLP) & 77.6 & 77.8 & \better{78.3} & 77.9 & \better{78.6} & \better{78.1} & 78.1 & 78.0 & 78.1 & 78.1 & 77.7 & \better{78.8} \\
RE (WLP) & 64.9 & \better{65.5} & \better{65.6} & 64.9 & \better{66.5} & \better{65.9} & 64.7 & 64.4 & 64.7 & 64.3 & 64.9 & \better{65.3} \\
RE (SemEval10) & 50.7 & \better{52.3} & \better{52.8} & \worse{49.6} & \better{52.9} & \better{52.8} & 61.7 & 61.9 & \worse{60.2} & \worse{59.2} & 62.1 & \worse{59.9} \\
ABSA & 63.5 & 63.4 & \worse{62.8} & \worse{59.8} & 63.5 & \worse{60.2} & 70.8 & \worse{68.9} & \better{71.4} & 70.4 & \worse{69.9} & \worse{69.6} \\
ORL & 38.2 & \worse{35.7} & 37.9 & \worse{36.1} & 38.6 & \better{41.0} & 44.5 & \better{45.8} & 44.2 & 44.8 & \better{45.1} & \better{46.6} \\
SRL (10k) & 68.8 & \better{69.6} & 68.9 & \better{70.7} & \better{71.3} & - & 78.7 & \better{79.4} & \better{79.5} & \better{79.6} & \better{79.8} & - \\
\bottomrule
\end{tabular}
}
\captionof{table}{Performance of pairwise multi-task learning with GloVe and BERT$_\text{base}$. \better{blue} indicates results better than STL, \worse{red} indicates worse, and black means almost the same (i.e., a difference within 0.5).
We show the performance after fine-tuning.
Dataset of source tasks POS, Consti., Dep. is PTB and dataset of NER is CoNLL-2003.
}
\vspace{-3mm}
\label{tab:mtl_pairwise}
\end{minipage}
\hfill
\begin{minipage}[b]{0.2\textwidth}
\centering
\includegraphics[width=1.0\columnwidth, clip, keepaspectratio]{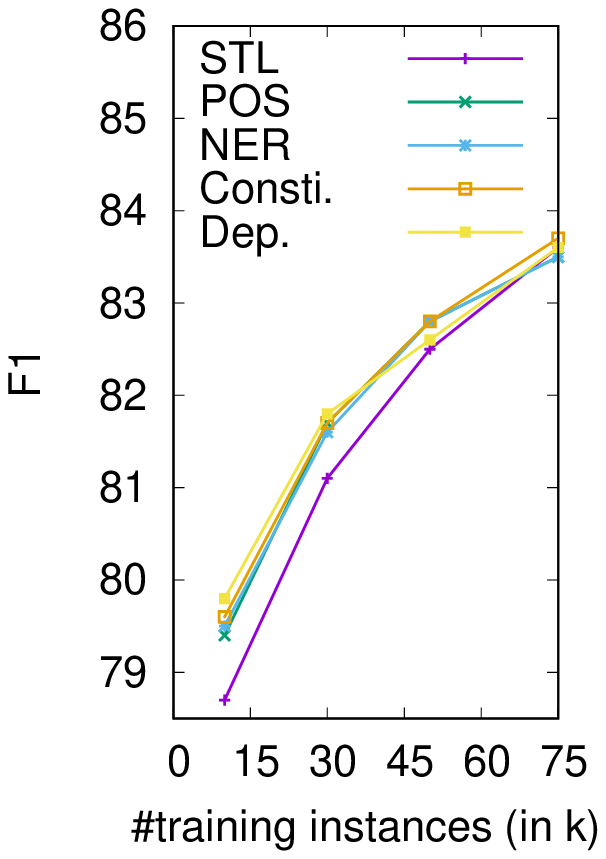}
\captionof{figure}{MTL Performance of SRL wrt. the data size.}
\vspace{-3mm}
\label{fig:srl_size}
\end{minipage}
\end{minipage}
\end{figure*}

We hypothesize that although language modeling pre-training is theoretically orthogonal to MTL \citep{swayamdipta-etal-2018-syntactic}, in practice their benefits tends to overlap.
To analyze these two factors separately, we start with a weak representation GloVe to study task relatedness, then move to BERT to demonstrate how much we can still improve with MTL given strong and contextualized representations.
As shown in Table~\ref{tab:mtl_pairwise} (GloVe), tasks are not equally useful to each other.
Notably, (1) for OpenIE and ORL, multi-task learning with SRL improves the performance significantly, while other tasks lead to less or no improvements.
(2) Dependency parsing and SRL are generic source tasks that are beneficial to most of the target tasks.
This unified \modelname makes it easy to perform MTL and decide beneficial source tasks.

\begin{table*}[tb]
\centering
\resizebox{\textwidth}{!}{%
\begin{tabular}{cccc|c@{\tightcol}c@{\tightcol}c|c@{\tightcol}c@{\tightcol}c|c@{\tightcol}c@{\tightcol}c|c@{\tightcol}c@{\tightcol}c}
\toprule
 & & & & \multicolumn{3}{c|}{\textbf{GloVe}} & \multicolumn{3}{c|}{\textbf{ELMo}} & \multicolumn{3}{c|}{\textbf{BERT$_\text{base}$}} & \multicolumn{3}{c}{\textbf{SpanBERT$_\text{base}$}} \\
\textbf{Category}  & \textbf{Task} & \textbf{Metric} & \textbf{Dataset}  & \textbf{STL} & \textbf{MTL} & \textbf{+FT} & \textbf{STL} & \textbf{MTL} & \textbf{+FT} & \textbf{STL} & \textbf{MTL} & \textbf{+FT} & \textbf{STL} & \textbf{MTL} & \textbf{+FT} \\ 
\midrule
\multirow{6}{*}{IE} & \multirow{2}{*}{NER} & \multirow{2}{*}{F$_1$} & CoNLL03 & 88.4 & \worse{86.2} & \worse{87.5} & 91.9 & 91.6 & 91.6 & 91.0 & \worse{88.6} & \worse{90.2} & 91.3 & \worse{90.4} & 91.2 \\
 & & & WLP & 77.6 & \worse{71.5} & \worse{76.5} & 79.2 & \worse{77.4} & \worse{78.2} & 78.1 & 78.2 & 78.5 & 77.9 & \better{78.6} & \better{78.5} \\
\cmidrule{2-16}
 & \multirow{2}{*}{RE} & \multirow{2}{*}{F$_1$} & SemEval10 & 50.7 & \worse{15.2} & \worse{33.0} & 61.8 & \worse{30.6} & \worse{42.9} & 61.7 & \worse{55.1} & \worse{59.8} & 62.1 & \worse{54.6} & 61.8 \\
 & & & WLP & 64.9 & \worse{38.5} & \worse{53.9} & 65.5 & \worse{52.0} & \worse{55.1} & 64.7 & \better{65.9} & \better{66.5} & 64.1 & \better{67.2} & \better{67.2} \\
\cmidrule{2-16}
 & Coref & Avg F$_1$ & OntoNotes & 56.3 & \worse{50.3} & \worse{53.0} & 62.2 & \better{62.9} & \better{63.3} & 66.2 & \worse{65.5} & 65.8 & 70.0 & \worse{68.9} & 69.7 \\
\cmidrule{2-16}
 & OpenIE & F$_1$ & OIE2016 & 28.3 & \worse{6.8} & \worse{19.6}  & 35.2 & \worse{30.0} & \worse{32.9} & 36.7 & 37.1 & \better{38.5} & 36.5 & \better{37.3} & \better{38.6}  \\ 
\midrule
\multicolumn{2}{c}{SRL} & F$_1$ & OntoNotes & 78.0 & 77.9 & \better{78.6}  & 82.4 & 82.3 & 82.4 & 83.3 & 82.9 & 83.4 & 83.1 & 83.3 & \better{83.8} \\ 
\midrule
\multirow{4}{*}{Parsing} & \multirow{2}{*}{Dep.} & \multirow{2}{*}{LAS} & PTB & 92.9 & 93.2 & \better{93.5} & 94.7 & 94.9 & 94.9 & 94.9 & 94.8 & 95.0 & 95.1 & 95.1 & 95.1 \\
 & & & OntoNotes & 90.4 & 90.5 & 90.5 & 92.3 & \better{93.2} & \better{92.8} & 94.1 & 93.8 & 94.0 & 94.2 & 94.1 & 94.2 \\
\cmidrule{2-16}
 & \multirow{2}{*}{Consti.} & \multirow{2}{*}{Evalb F$_1$} & PTB & 93.4 & - & 93.8 & 95.3 & - & 95.3 & 95.5 & - & 95.2 & 95.8 & - & 95.5 \\
 & & & OntoNotes & 91.0 & - & \better{91.5} & 93.2 & - & \better{93.7} & 93.6 & - & 93.8 & 94.3 & - & 94.2 \\ 
\midrule
\multirow{2}{*}{Sentiment} & ABSA & F$_1$ & SemEval14 & 63.5 & \worse{48.5} & \worse{59.0} & 69.2 & \worse{57.0} & \worse{59.0} & 70.8 & \worse{63.1} & \worse{67.0} & 70.0 & \worse{63.5} & \worse{69.5} \\
 & ORL & F$_1$ & MPQA 3.0 & 38.2 & \worse{18.4} & \worse{31.6} & 42.9 & \worse{24.7} & \worse{32.4} & 44.5 & \worse{38.1} & \better{45.6} & 45.2 & \worse{40.2} & \better{47.5} \\ 
\midrule
\multicolumn{2}{c}{\multirow{2}{*}{POS}} & \multirow{2}{*}{Accuracy} & PTB & 96.8 & 96.8 & 96.8 & 97.7 & 97.7 & 97.8 & 97.6 & 97.3 & 97.3 & 97.6 & 97.6 & 97.6 \\
\multicolumn{2}{c}{} & & OntoNotes & 97.0 & 97.0 & 97.1 & 98.2 & 98.2 & 98.3 & 97.7 & 97.8 & 97.8 & 98.3 & 98.3 & 98.3 \\
\bottomrule
\end{tabular}
}
\caption{Comparison between STL and MTL+fine-tuning across all tasks.
\better{blue} indicates results better than STL, \worse{red} indicates worse, and black means almost the same (i.e., a difference within 0.5).
Constituency parsing requires more memory than other tasks so we restrict its span length to 10 in MTL, and thus do not report results.
}
\vspace{-3mm}
\label{tab:mtl}
\end{table*}

\begin{figure}[tb]
\centering
\begin{subfigure}{0.56\columnwidth}
\centering
\includegraphics[width=1.0\linewidth]{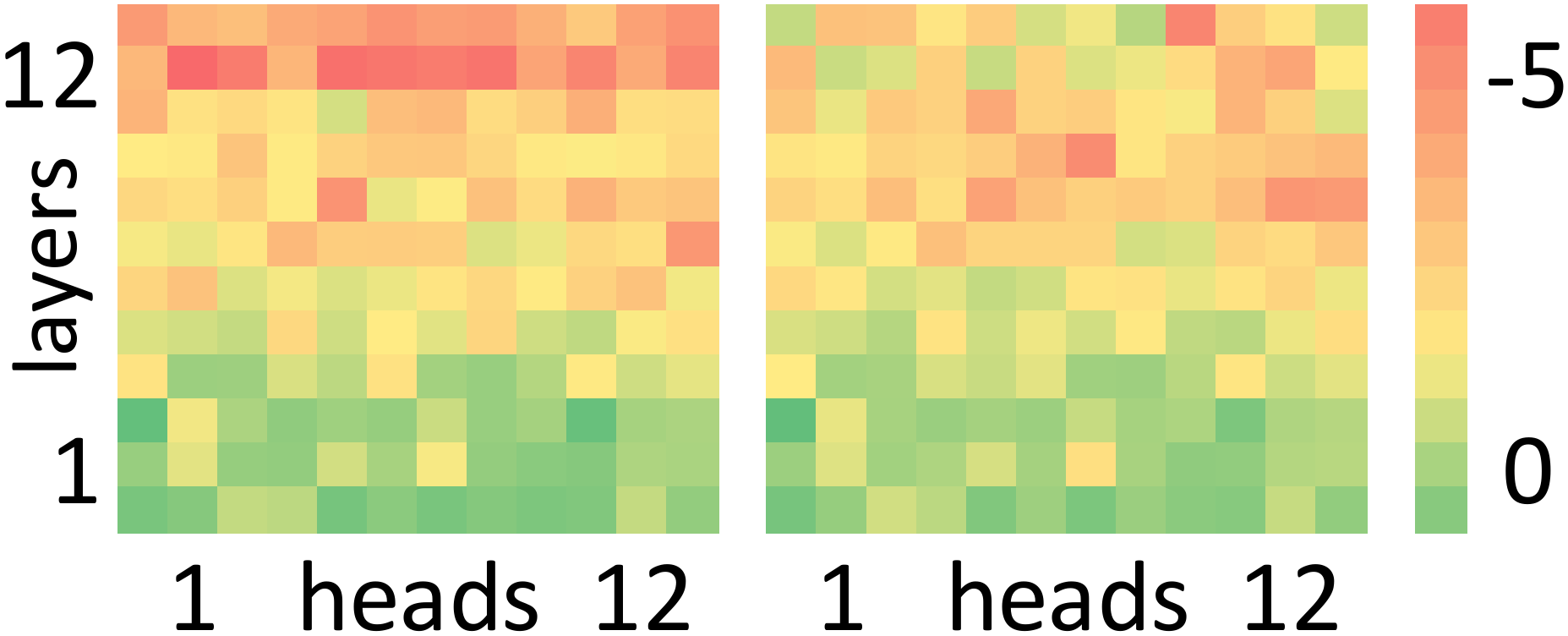}
\caption{Attention similarity between OpenIE/POS (left), and between OpenIE/SRL (right) for all heads.}
\label{fig:attn_vis}
\end{subfigure}
\hfill
\begin{subfigure}{0.42\columnwidth}
\centering
\includegraphics[width=1.0\linewidth]{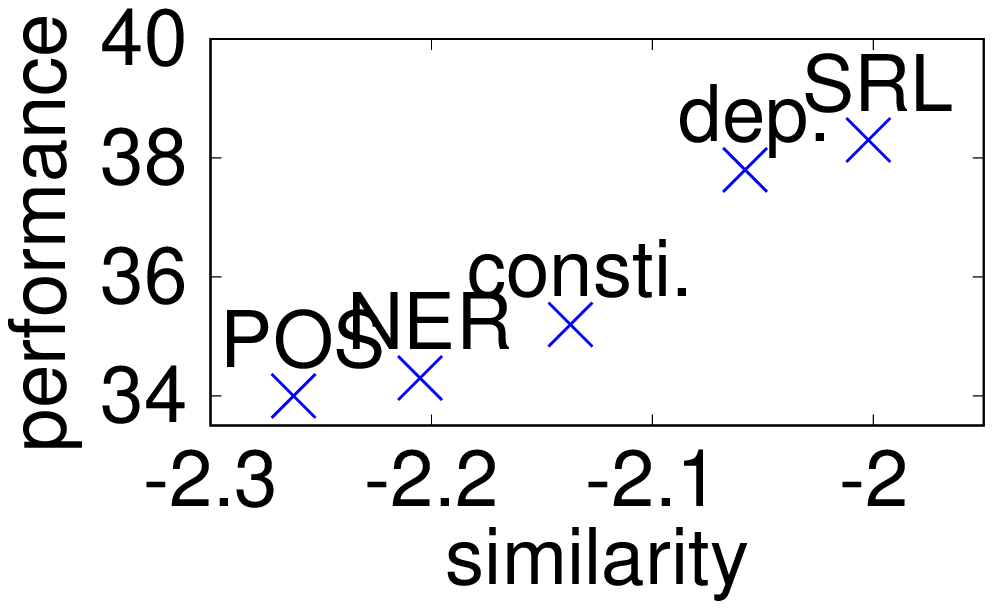}
\caption{Correlation between attention similarity and MTL performance.}
\label{fig:est}
\end{subfigure}
\caption{Attention-based task relatedness analysis.}
\vspace{-3mm}
\end{figure}

Next, we demonstrate that our framework also provides a platform for analysis of similarities and differences between different tasks.
Inspired by the intuition that the attention coefficients are somewhat indicative of a model's internal focus \citep{li-2016-understand,vig-2019-bertvis,clark-2019-bert}, we hypothesize that the similarity or difference between attention mechanisms may be correlated with similarity between tasks, or even the success or failure of MTL.
To test this hypothesis, we extract the attention maps of two BERT-based \modelname models (trained on a source $t^\prime$ and a target task $t$ separately) over sentences $\mathcal{X}_t$ from the target task, and compute their similarity using the Frobenius norm:
\begin{equation*}
\text{sim}_k(t, t^\prime) = -\frac{1}{|\mathcal{X}_t|} \sum_{\mathbf{x} \in \mathcal{X}_t}{\left\Vert A_k^t(\mathbf{x}) - A_k^{t^\prime}(\mathbf{x})\right\Vert_F},
\end{equation*}
where $A_k^t(\mathbf{x})$ is the attention map extracted from the $k$-th head by running the model trained from task $t$ on sentence $\mathbf{x}$.
We select OpenIE as the target task because it shows the largest performance variation when paired with different source tasks (34.0 - 38.8) in Table~\ref{tab:mtl_pairwise}.
We visualize the attention similarity of all heads in BERT (12 layers $\times$ 12 heads) between two mutually harmful tasks (OpenIE/POS on the left) and between two mutually helpful tasks (OpenIE/SRL on the right) in \autoref{fig:attn_vis}.
A common trend is that heads in higher layers exhibit more divergence, probably because they are closer to the prediction layer, thus easier to be affected by the end task.
Overall, it can be seen that OpenIE/POS has much more attention divergence than OpenIE/SRL.
A notable difference is that almost \textit{all} heads in the last two layers of the OpenIE/POS models differ significantly, while \textit{some} heads in the last two layers of the OpenIE/SRL models still behave similarly, providing evidence that failure of MTL can be attributed to the fact that dissimilar tasks requires different attention patterns.
We further compute average attention similarities for all source tasks in \autoref{fig:est}, and we can see that there is a strong correlation (Pearson correlation of 0.97) between the attentions similarity and the performance of pairwise MTL, supporting our hypothesis that attention pattern similarities can be used to predict improvements of MTL.

\paragraph{MTL under Different Settings}
We analyze how token representations and sizes of the target dataset affect the performance of MTL.
Comparing BERT and GloVe in Table~\ref{tab:mtl_pairwise}, the improvements become smaller or vanish as the token representation becomes stronger, e.g., improvement on OpenIE with SRL reduces from 5.8 to 1.6.
This is expected because both large-scale pre-training and MTL aim to learn general representations and their benefits tend to overlap in practice.
Interestingly, some helpful source tasks become harmful when we shift from GloVe to BERT, such as OpenIE paired with POS.
We conjecture that the gains of MTL might have already been achieved by BERT, but the task-specific characteristics of POS hurt the performance of OpenIE.
We did not observe many tasks benefitting from MTL for the GloVe-based model in \autoref{tab:mtl} because it is trained on \textit{all} tasks (instead of \textit{two}), which is beyond its limited model capacity.
The improvements of MTL shrink as the size of the SRL datasets increases, as shown in \autoref{fig:srl_size}, indicating that MTL is useful when the target data is sparse.

\paragraph{Time Complexity Analysis}
Time complexities of span and relation prediction are $\mathcal{O}(l \cdot n)$ and $\mathcal{O}(K^2)=\mathcal{O}(\tau^2 \cdot n^2)$ respectively for a sentence of $n$ tokens (\autoref{sec:spanrelationmodel}).
The time complexity of BERT is $\mathcal{O}(L \cdot n^2)$, dominated by its $L$ self-attention layers.
Since the pruning threshold $\tau$ is usually less than 1, the computational overhead introduced by the span-relation output layer is much less than BERT.
In practice, we observe that the training/testing time is mainly spent by BERT. For SRL, one of the most computation-intensive tasks with long spans and dense span/relation annotations, 85.5\% of the time is spent by BERT. For POS, a less heavy task, the time spent by BERT increases to 98.5\%.
Another option for span prediction is to formulate it as a sequence labeling task, as in previous works \citep{lample:16:ner,he-etal-2017-deep}, where time complexity is $\mathcal{O}(n)$. Although slower than token-based labeling models, span-based models offer the advantages of being able to model overlapping spans and use span-level information for label prediction \citep{lee:17:e2ecoref}.

\section{Related Work}\label{sec:related}

\paragraph{General Architectures for NLP}
There has been a rising interest in developing general architectures for different NLP tasks, with the most prominent examples being sequence labeling framework \citep{collobert:11:nlpscratch,ma:16:lstmcnncrf} used for tagging tasks and sequence-to-sequence framework \citep{sutskever:14:seq2seq} used for generation tasks.
Moreover, researchers typically pick related tasks, motivated by either linguistic insights or empirical results, and create a general framework to perform MTL, several of which are summarized in \autoref{tab:related_work}.
For example, \citet{swayamdipta-etal-2018-syntactic} and \citet{strubell-etal-2018-linguistically} use constituency and dependency parsing to improve SRL.
\citet{luan:18:mtlie,luan-etal-2019-general,wadden-etal-2019-entity} use a span-based model to jointly solve three information-extraction-related tasks (NER, RE, and Coref.).
\citet{li-2019-mrcner} formulate both nested NER and flat NER as a machine reading comprehension task.
Compared to existing works, we aim to create an output representation that can solve \emph{nearly every} natural language analysis task in one fell swoop, allowing us to cover a far broader range of tasks with a single model.

In addition, NLP has seen a recent burgeoning of contextualized representations pre-trained on large corpora (e.g., ELMo \citep{peters-etal-2018-deep} and BERT \citep{devlin-etal-2019-bert}).
These methods focus on learning generic \emph{input} representations, but are agnostic to the \emph{output} representation, requiring different predictors  for different tasks. In contrast, we present a methodology to formulate the output of different tasks in a unified format.
Thus our work is orthogonal to those on contextualized embeddings.
Indeed, in \autoref{sec:exp_mtl}, we demonstrate that the \modelname model can benefit from stronger contextualized representation models, and even provide a testbed for their use in natural language analysis.

\paragraph{Benchmarks for Evaluating Natural Language Understanding} Due to the rapid development of NLP models, large-scale benchmarks, such as SentEval \citep{conneau-kiela-2018-senteval}, GLUE \citep{wang:19:glue}, and SuperGLUE \citep{wang:19:superglue} have been proposed to facilitate fast and holistic evaluation of models' understanding ability.
They mainly focus on sentence-level tasks, such as natural language inference, while our \benchmarkname benchmark focuses on token/phrase-level analysis tasks with diverse coverage of different linguistic structures.
New tasks and datasets can be conveniently added to our benchmark as long as they are in the BRAT standoff format, which is one of the most commonly used data format in the NLP community, e.g., it has been used in the BioNLP shared tasks \citep{kim:2009:bionlp} and the Universal Dependency project \citep{mcdonald-etal-2013-universal}.

\section{Conclusion}
We provide the simple insight that a large number of natural language analysis tasks can be represented in a single format consisting of spans and relations between spans.
As a result, these tasks can be solved in a single modeling framework that first extracts spans and predicts their labels, then predicts relations between spans.
We attempted 10 tasks with this \modelname model and show that this generic task-independent model can achieve competitive performance as state-of-the-art methods tailored for each tasks.
We merge 8 datasets into our \benchmarkname benchmark for evaluating future models for natural language analysis.
Future directions include (1) devising hierarchical span representations that can handle spans of different length and diverse content more effectively and efficiently; (2) robust multitask learning or meta-learning algorithms that can reconcile very different tasks.

\section*{Acknowledgments}
This work was supported by gifts from Bosch Research. We would like to thank Hiroaki Hayashi, Bohan Li, Pengcheng Yin, Hao Zhu, Paul Michel, and Antonios Anastasopoulos for their insightful comments and suggestions.

\bibliography{acl2020}
\bibliographystyle{acl_natbib}

\newpage
\appendix

\section{Detailed Explanations of 10 Tasks}\label{ap:tasks}

\begin{itemize}[leftmargin=10pt]
\item \textbf{Span-oriented Tasks (\subref{tab:demo_span})}
\begin{itemize}[leftmargin=8pt]
\item \textbf{Named Entity Recognition} \citep{sang:03:conll2003}
NER is traditionally considered as a sequence labeling task. We model named entities as spans over one or more tokens.
\item \textbf{Constituency Parsing} \citep{collins-1997-three} Constituency parsing aims to produce a syntactic parse tree for each sentence. Each node in the tree is an individual span associated with a constituent label, and spans are nested.
\item \textbf{Part-of-speech Tagging} \citep{ratnaparkhi-1996-maximum,toutanova-etal-2003-feature} POS tagging is another sequence labeling task, where every single token is an individual span with a POS tag.
\item \textbf{Aspect-based Sentiment Analysis} \citep{pontiki:14:semeval2014} ABSA is a task that consists of identifying certain spans as aspect terms and predicting their associated sentiments.
\end{itemize}
\item \textbf{Relation-oriented Tasks (\subref{tab:demo_rel})}
\begin{itemize}[leftmargin=8pt]
\item \textbf{Relation Extraction} \citep{hendrickx:10:semeval2010} RE concerns the relation between two entities.
\item \textbf{Coreference} \citep{pradhan:12:conll2012} Coreference resolution is to link named, nominal, and pronominal mentions that refer to the same concept, within or beyond a single sentence. 
\item \textbf{Semantic Role Labeling} \citep{gildea:2002:srl} SRL aims to identify arguments of a predicate (verb or noun) and classify them with semantic roles in relation to the predicate.
\item \textbf{Open Information Extraction} \citep{banko:07:openie,niklaus-etal-2018-survey} In contrast to the fixed relation types in RE, OpenIE aims to extract open-domain predicates and their arguments (usually subjects and objects) from a sentence.
\item \textbf{Dependency Parsing} \citep{kubler2009dependency} Spans are single-word tokens and a relation links a word to its syntactic parent with the corresponding dependency type. 
\item \textbf{Opinion Role Labeling} \citep{yang:13:fgoe} 
ORL detects spans that are opinion expressions, as well as holders and targets related to these opinions.
\end{itemize}
\end{itemize}

\section{Results of BERT Large Model}\label{ap:bertlarge}

\autoref{tab:stl} shows the performance of single-task learning with different token representations.
BERT$_\text{large}$ achieves the best performance on most of the tasks.

\begin{table*}[t]
\centering
\resizebox{\textwidth}{!}{
\begin{tabular}{cccc|ccccc}
\toprule
\textbf{Category} & \textbf{Task} & \textbf{Metric} & \textbf{Dataset} & \textbf{GloVe} & \textbf{ELMo} & \textbf{BERT$_\text{base}$} & \textbf{SpanBERT$_\text{base}$} & \textbf{BERT$_\text{large}$} \\
\midrule
\multirow{6}{*}{IE} & \multirow{2}{*}{NER} & \multirow{2}{*}{F$_1$} & CoNLL03 & 88.4 & 91.9 & 91.0 & 91.3 & 90.9 \\
 & & & WLP & 77.6 & 79.2 & 78.1  & 77.9 & 78.3 \\
\cmidrule{2-9}
 & \multirow{2}{*}{RE} & \multirow{2}{*}{F$_1$} & SemEval10 & 50.7 & 61.8 & 61.7 & 62.1 & 64.7 \\
 & & & WLP & 64.9 & 65.5 & 64.7 & 64.1 & 65.1 \\
\cmidrule{2-9}
 & \multirow{1}{*}{Coref} & Avg F$_1$ & OntoNotes & 56.3 & 62.2 & 66.3 & 70.0 & - \\
\cmidrule{2-9}
 & \multirow{1}{*}{OpenIE} & F$_1$ & OIE2016 & 28.3 & 35.2 & 36.7 & 36.5 & 36.5 \\
\midrule
\multicolumn{2}{c}{SRL} & F$_1$ & OntoNotes & 78.0 & 82.4 & 83.3 & 83.1 & 84.4 \\
\midrule
\multirow{4}{*}{Parsing} & \multirow{2}{*}{Dep.} & \multirow{2}{*}{LAS} & PTB & 92.9 & 94.7 & 94.9 & 95.1 & 95.3 \\
 & & & OntoNotes & 90.4 & 92.3 & 94.1 & 94.2 & 94.5 \\
\cmidrule{2-9}
 & \multirow{2}{*}{Consti.} & \multirow{2}{*}{Evalb F$_1$} & PTB & 93.4 & 95.3 & 95.5 & 95.8 & 95.8 \\
 & & & OntoNotes & 91.0 & 93.2 & 93.6 & 94.3 & 93.9 \\
\midrule
\multirow{2}{*}{Sentiment} & \multirow{1}{*}{ABSA} & F$_1$ & SemEval14 & 63.5 & 69.2 & 70.8 & 70.0 & 73.8 \\
\cmidrule{2-9}
 & \multirow{1}{*}{ORL} & F$_1$ & MPQA 3.0 & 38.2 & 42.9 & 44.5 & 45.2 & 47.1 \\
\midrule
\multicolumn{2}{c}{\multirow{2}{*}{POS}} & \multirow{2}{*}{Accuracy} & PTB & 96.8 & 97.7 & 97.6 & 97.6 & 97.4 \\
 & & & OntoNotes & 97.0 & 98.2 & 97.7 & 98.3 & 97.9 \\
\bottomrule
\end{tabular}
}
\caption{Single-task learning performance of the \modelname model with different token representations. BERT$_\text{large}$ requires a large amount of memory so we cannot feed the entire document to the model in coreference resolution.
}
\label{tab:stl}
\end{table*}

\begin{table*}[t]
\small
\centering
\begin{tabular}{ccccccccccc}
\toprule
 & \multicolumn{4}{c}{\textbf{Information Extraction}} & \multirow{2}{*}{\textbf{POS}} & \multicolumn{2}{c}{\textbf{Parsing}} & \multirow{2}{*}{\textbf{SRL}} & \multicolumn{2}{c}{\textbf{Sentiment}} \\
\cline{2-5} 
\cline{7-8}
\cline{10-11}
 & NER & RE & Coref. & OpenIE  & & Dep. & Consti. & & ABSA & ORL\\
\toprule
max span length $l$ & 10 & 5 & 10 & 30 & 1 & 1 & - & 30 & 10 & 30 \\
pruning ratio $\tau$ & - & 5 & 0.4 & 0.8 & - & 1.0 & - & 1.0 & - & 0.3 \\
\bottomrule
\end{tabular}
\caption{Task-specific hyperparameters. Span-oriented tasks do not need pruning ratio.}
\label{tab:hyperparam}
\end{table*}

\section{Task-specific Hyperparameters}\label{ap:hyper}

As shown in \autoref{tab:hyperparam}, a larger maximum span length is used for tasks with longer spans (e.g., OpenIE), and a larger pruning ratio is used for tasks with more spans (e.g., SRL).
Constituency parsing does not have span length limit because spans can be as long as the entire sentence.
Since relation extraction aims to extract exactly two entities and their relation from a sentence, we keep pruning ratio fixed (top 5 spans in this case) regardless of the length of the sentence.

\end{document}